%% file: main.tex
\documentclass[11pt,a4paper]{article}
\usepackage[dvipsnames,table]{xcolor}
\usepackage[hyperref]{emnlp2020}
\usepackage{times}

\usepackage{latexsym}

\usepackage{microtype}

\usepackage{booktabs}
\usepackage{amssymb}
\usepackage{xspace}

\usepackage{bbm}
\usepackage{amsmath}
\usepackage{todonotes}
\usepackage{url}
\usepackage{comment}
\usepackage{subcaption}
\usepackage{cleveref}
\usepackage{float}

\usepackage{pifont}%
\newcommand{\cmark}{\ding{51}}%
\newcommand{\xmark}{\ding{55}}%

\aclfinalcopy %
\def\aclpaperid{1528} %

\usepackage[dvipsnames,table]{xcolor}
\newcommand{\ourapproach}{{\sc EaE}} 

\newcommand{\relic}{{\sc relic}}
\newcommand{\lama}{{\sc lama}}
\newcommand{\triviaqa}{TriviaQA}
\newcommand{\bert}{\textsc{Bert}}

\newcommand{\tfive}{T5}

\newcommand{\blank}{\textsc{[mask]}}
\newcommand{\ctx}{\mathbf{x}}

\newcommand{\knowbert}{\textsc{KnowBert}\xspace}
\newcommand{\ernie}{\textsc{Ernie}\xspace}
\newcommand{\bertmm}{BERT-MM}

\newcolumntype{g}{>{\color{gray}}c}

\title{Entities as Experts: Sparse Memory Access with Entity Supervision}

\author{Thibault F\'evry~\thanks{~~Work done during Google AI residency.} \\ \normalsize{\texttt{thibaultfevry@gmail.com}} \And Livio Baldini Soares \\ \normalsize{\texttt{liviobs@google.com}} \And Nicholas FitzGerald \\ \normalsize{\texttt{nfitz@google.com}}
\AND Eunsol Choi\thanks{~~Work done at Google Research.} \\ The University of Austin at Texas \\ \normalsize{\texttt{eunsol@cs.utexas.edu}} \And Tom Kwiatkowski \\ \normalsize{\texttt{tomkwiat@google.com}} \AND Google Research   }

\date{}

\begin{document}
\maketitle
\begin{abstract}
We focus on the problem of capturing declarative knowledge about entities in the learned parameters of a language model.
We introduce a new model---Entities as Experts (\ourapproach{})---that can access distinct memories of the entities mentioned in a piece of text.
Unlike previous efforts to integrate entity knowledge into sequence models, \ourapproach{}'s entity representations are learned directly from text.
We show that \ourapproach{}'s learned representations capture sufficient knowledge to answer TriviaQA questions such as {\it ``Which Dr. Who villain has been played by Roger Delgado, Anthony Ainley, Eric Roberts?''}, outperforming an encoder-generator Transformer model with $10\times$ the parameters. 
According to the {\sc lama} knowledge probes, \ourapproach{} contains more factual knowledge than a similarly sized {\sc Bert}, as well as previous approaches that integrate external sources of entity knowledge.
Because \ourapproach{} associates parameters with specific entities, it only needs to access a fraction of its parameters at inference time, and we show that the correct identification and representation of entities is
essential to \ourapproach{}'s performance.
\end{abstract}

\input{introduction.tex}

\input{approach.tex}

\input{background.tex}

\input{our_models.tex}

\input{knowledge_probing_tasks_and_results.tex}

\input{analysis.tex}

\input{entity_rep_comparison}

\input{conclusion}

\section*{Acknowledgments}
We thank many members of Google Research for helpful feedback and discussions, especially Kenton Lee, Dipanjan Das, Pat Verga and William Cohen. We also thank Adam Roberts and Sewon Min for sharing their system outputs.

\bibliography{eae}
\bibliographystyle{acl_natbib}

\appendix
\clearpage
\input{appendix.tex}

\end{document}

%% file: introduction.tex
\section{Introduction}\label{sec:intro}
Neural network sequence models, pre-trained as language models, have recently revolutionized text understanding \cite{dai2015semi, elmo, ulmfit, bert}, and recent work has suggested that they could take the place of curated knowledge bases or textual corpora for tasks such as question answering  \cite{language_as_kb, t5_openqa}.

In this paper, we focus on developing neural sequence models that capture the knowledge required to answer questions about real world entities.
To this end, we introduce a new model architecture that can access distinct and independent representations of the entities mentioned in text.
Unlike other efforts to inject entity specific knowledge into sequence models \cite{knowbert, ernie, ebert} our model learns entity representations from text along with all the other model parameters.
\begin{figure}[t]
    \centering
    \includegraphics[width=0.48\textwidth]{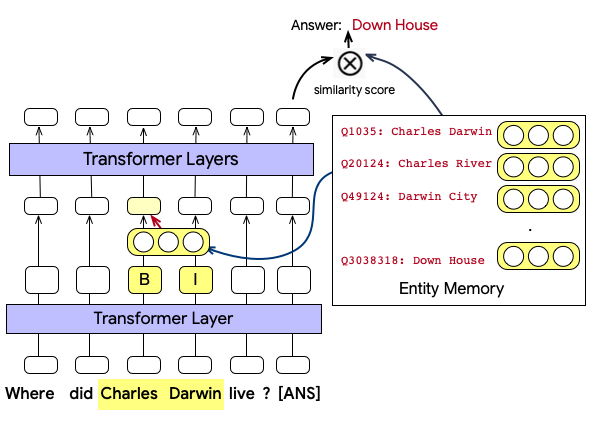}
    \vspace*{-10mm}
    \caption{Our model with an entity memory, applied to the open domain QA task. The red arrows shows the integration of the entity and token representations.}
    \vspace*{-4mm}
    \label{fig:teaser}
\end{figure}
We call our model Entities as Experts (\ourapproach{}), since it divides the parameter space according to entity identity.
This name also reflects \ourapproach{}'s similarities with the Massive Mixture of Experts \cite{mixtureofexperts}, as well as other work that integrates learned memory stores into sequence models \cite{memory_networks, largepkm}.

To understand the motivation for distinct and independent entity representations, consider Figure~\ref{fig:teaser}.
A traditional Transformer \cite{transformer} needs to build an internal representation of Charles Darwin from the words ``Charles'' and ``Darwin'', both of which can also refer to different entities such as the Charles River, or Darwin City.
Conversely, \ourapproach{} can access a dedicated representation of ``Charles Darwin'', which is a memory of all of the contexts in which this entity has previously been mentioned. This memory can also be accessed for other mentions of Darwin, such as ``Charles Robert Darwin'' or ``the father of natural selection''. Retrieving and re-integrating this memory makes it easier for \ourapproach{} to find the answer.

We train \ourapproach{} to predict masked-out spans in English Wikipedia text \cite{bert}; to only access memories for entity mention spans; and to access the correct memory for each entity mention.
Mention span supervision comes from an existing mention detector, and entity identity supervision comes from Wikipedia hyperlinks.
By associating memories with specific entities, \ourapproach{} can learn to access them sparsely. The memory is only accessed for spans that mention entities, and only the mentioned memories need to be retrieved.

We evaluate \ourapproach{}'s ability to capture declarative knowledge using the \lama{} knowledge probes introduced by \newcite{language_as_kb}, as well as the open-domain variants of the TriviaQA and Web\-Questions question answering tasks \cite{triviaqa, webquestions}.
On both tasks, \ourapproach{} outperforms related approaches with many more parameters. An in-depth analysis of \ourapproach{}'s predictions on TriviaQA shows that the correct identification and reintegration of entity representations is essential for \ourapproach{}'s performance.

We further demonstrate that \ourapproach{}'s learned entity representations are better than the pre-trained embeddings used by \citet{ernie, knowbert} at knowledge probing tasks and the TACRED relation extraction task~\cite{tacred, tacred-revisited}.
We show that training \ourapproach{} to focus on entities is better than imbuing a similar-sized network with an unconstrained memory store, and explain how \ourapproach{} can outperform much larger sequence models while only accessing a small proportion of its parameters at inference time.

%% file: approach.tex
\begin{figure*}
\vspace{-10pt}
    \includegraphics[width=\linewidth]{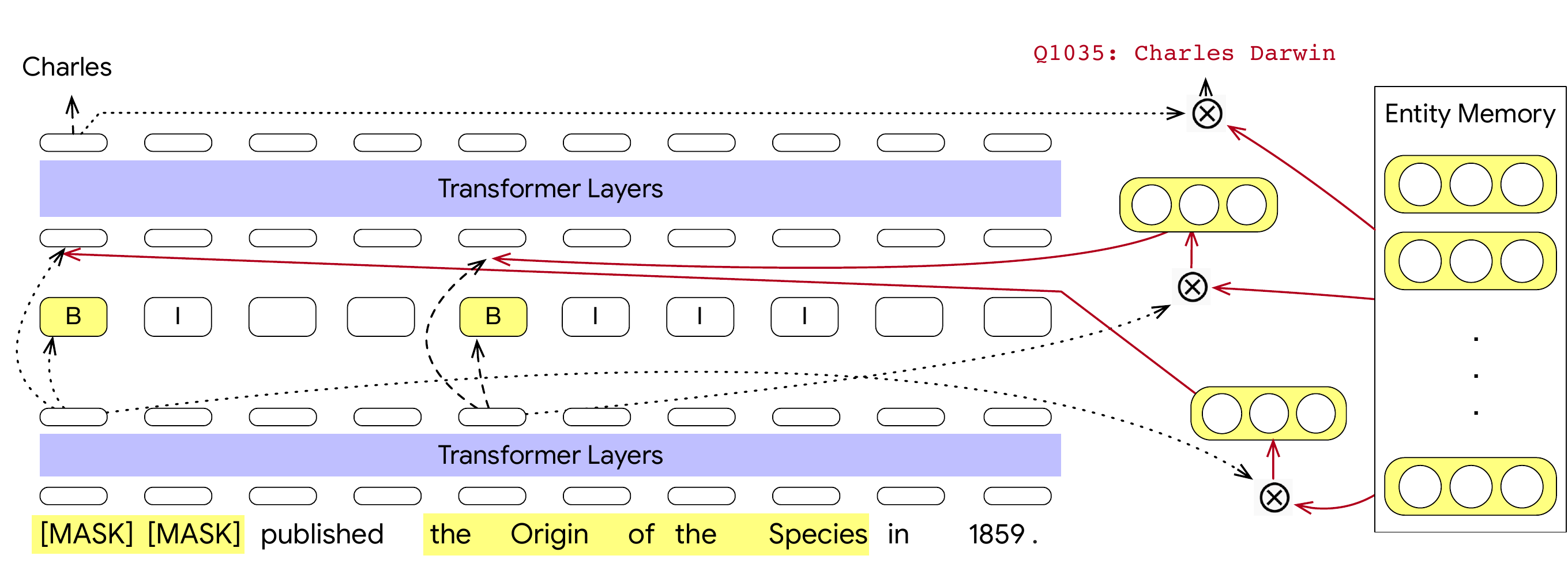}
    \caption{The \textbf{E}ntities \textbf{a}s \textbf{E}xperts model: the initial transformer layer output is used (i) to predict mention boundaries, (ii) to retrieve entity embeddings from entity memory, and (iii) to construct input to the next transformer layer, augmented with the retrieved entity embeddings of (ii). The final transformer block output is connected to task specific heads: token prediction and entity prediction. The entity retrieval after the first transformer layer (ii) is also supervised with an entity linking objective during pre-training.}
    \label{fig:architecture}
\end{figure*}

\section{Approach}\label{sec:model}

Let $\mathcal{E} = \{e_1 \dots e_N\}$ be a predefined set of entities, and let $\mathcal{V} = \{\blank, w_1 \dots w_M\}$ be a vocabulary of tokens. A \emph{context} $\ctx = [x_0 \dots  x_L]$ is a sequence of tokens $x_i \in \mathcal{V}$. Each context comes with the list of the \emph{mentions} it contains, $\mathbf{m} = [m_0 \dots m_M]$, where each mention $m_i = (e_{m_i}, s_{m_i}, t_{m_i})$ is defined by its linked entity $e_{m_i}$, start token index $s_{m_i}$ and end token index $t_{m_i}$. Entity mentions might not be linked to a specific entity in  $\mathcal{E}$, thus $e_{m_i} \in \mathcal{E} \cup e_\varnothing$, where $e_\varnothing$ refers to the null entity.

\subsection{Model Architecture}
The basic model architecture follows the Transformer~\cite{transformer}, interleaved with our entity memory layer. Our model has two embedding matrices -- token and entity embeddings. Figure~\ref{fig:architecture} illustrates our model. Our model is:

\vspace{-3mm}
{\small
\begin{flalign*}
&\mathbf{X}^0 = \mathtt{TokenEmbed}(\ctx) \\
&\mathbf{X}^1 = \mathtt{Transformer}(\mathbf{X}^0, \mathtt{num\_layers}=l_0)\\
&\mathbf{X}^2 =\mathtt{EntityMemory}(\mathbf{X}^1)\\
&\mathbf{X}^3 = \mathtt{LayerNorm}(\mathbf{X}^2+\mathbf{X}^1)\\
&\mathbf{X}^4 = \mathtt{Transformer}(\mathbf{X}^3,\mathtt{num\_layers}=l_1)\\
&\mathbf{X}^5 = \mathtt{TaskSpecificHeads}(\mathbf{X}^4)
\end{flalign*}}%
\nolinebreak{}
\nolinebreak 
The entity memory layer constructs an entity embedding $E_{m_i}$ for each mention $m_i$. The output of the entity memory layer and preceding transformer layer are summed, normalized, then processed by additional transformer layers. Throughout this work we use $l_0=4$ and $l_1=8$.

\paragraph{Entity Memory Layer}
Let $\mathbf{E}$ be a matrix of learned entity embeddings of shape $(N, d_{ent})$. $\mathtt{EntEmbed}(e_i)$ maps an entity $e_i$ to its row in $\mathbf{E}$. The entity memory layer takes the output sequence from the preceding transformer layer ($\mathbf{X}^l$) and outputs a new sequence ($\mathbf{X}^{l+1}$), sparsely populated with entity representations. For each entity mention $m_i$, the output sequence has an entity representation, a projection of the weighted sum of entity embeddings in $\mathbf{E}$, at position $s_{m_i}$. 
\begin{equation}
{x^{l+1}_i} = \mathbf{W}_{b} E_{m_k} \quad \mathtt{if }\hspace{0.1in} i= s_{m_k}  
\end{equation}
where $\mathbf{W}_{b}$ maps the entity representation $E_{m_k}$ to the dimension of $x^l_i$.

We now describe how to generate $E_{m_i}$ for each mention $m_i$. First, we generate a pseudo entity embedding $h_{m_i}$ based on the mention's span representation $[x^l_{s_{m_i}} || x^l_{t_{m_i}}]$, a concatenation of its start and tail representations. 
\begin{equation}
h_{m_i} = \mathbf{W_f} [x^l_{s_{m_i}} || x^l_{t_{m_i}}] \label{eq:pseudo}
\end{equation}
where $\mathbf{W_f}$ is of shape $({d_{ent} , 2 \cdot d_{emb}})$, where $d_{emb}$ is the dimension of $\mathbf{X^1}$.

We find the $k$ nearest entity embeddings of $h_{m_i}$ from $\mathbf{E}$ by computing the dot product, and $E_{m_i}$ is a weighted sum of them. More formally:
\begin{align*}
E_{m_i} &= \sum_{e_j \in \mathtt{topK}(\mathcal{E}, h_{m_i},k)} \alpha_j \cdot (\mathtt{EntEmbed}(e_j)) \label{eq:top_k} \\
\alpha_j &= \frac{\exp(\mathtt{EntEmbed}(e_j) \cdot h_{m_i})}{\sum_{e \in \mathtt{topK}(\mathcal{E}, h_{m_i},k)}{\exp(\mathtt{EntEmbed}(e) \cdot h_{m_i}})}
\end{align*}
Where $\mathtt{topK}(\mathcal{E}, h_{m_i},k)$ returns the $k$ entities that yield the highest score $\mathtt{EntEmbed}(e_j)\cdot h_{m_i}$. We use $k=N$ to train and use $k=100$ at inference (see Section~\ref{sec:eae_model_description} and \ref{sec:top_k}).

The entity memory layer can be applied to any sequence output without loss of generality. We apply it to the output of the first Transformer. 

\paragraph{Task-Specific Heads}
The final transformer layer can be connected to multiple task specific heads. In our experiments, we introduce two heads: \texttt{TokenPred} and \texttt{EntityPred}.

The \texttt{TokenPred} head predicts masked tokens for a cloze task. Each masked token's final representation $x^4_i$ is fed to an output softmax over the token vocabulary, as in \textsc{Bert}.

The \texttt{EntityPred} head predicts entity ids for each entity mention span (i.e., entity linking). We build the pseudo entity embedding ($h_{m_i}$) from the last sequence output ($X^4$). Then, the model predicts the entity whose embedding in $\mathbf{E}$ is the closest to the pseudo entity embedding.

\paragraph{Inference-time Mention Detection}
We introduce a mention detection layer to avoid dependence at inference on an external mention detector.
The mention detection layer applies a \textsc{bio}\footnote{In a \textsc{bio} encoding, each token is classified as being the \textsc{B}eginning, \textsc{I}nside, or \textsc{O}utside of a mention.} classifier to the first transformer block's output. 
We decode the entire \textsc{bio} sequence, ensuring that inconsistent sequences are disallowed. We use inferred mention spans at inference for all our experiments.

\subsection{Training}\label{sec:training}

\subsubsection{Data and Preprocessing}\label{sec:data}
We assume access to a corpus $\mathcal{D} = \{(\ctx_i, \mathbf{m_i})\}$, where all entity mentions are detected but not necessarily all linked to entities.
We use English Wikipedia as our corpus, with a vocabulary of 1m entities.
Entity links come from hyperlinks, leading to 32m 128 byte contexts containing 17m entity links.
Non-overlapping entity mention boundaries come from hyperlinks and the Google Cloud Natural Language  API\footnote{\url{https://cloud.google.com/natural-language/docs/basics\#entity\_analysis}} leading to 140m mentions.

We remove 20\% of randomly chosen entity mentions (all tokens in the mention boundaries are replaced with \blank) to support a masked language modeling objective. See Appendix~\ref{sec:pretraining} for full details of our pre-processing and pre-training hyper-parameters.

\subsubsection{Learning Objective}\label{sec:objective}
The pre-training objective is the sum of (1) a mention boundary detection loss, (2) an entity linking loss, and (3) a masked language modeling loss. 

\paragraph{Mention Detection}
The \textsc{bio} classification of tokens is supervised with a cross-entropy loss over the labels.
Assuming the mention boundaries are complete, we apply this supervision to all tokens.

\paragraph{Entity Linking} 
We use the hyperlinked entities $e_{m_i}$ to supervise entity memory assess. For each hyperlinked mention $m_i=(e_{m_i}, s_{m_i},t_{m_i})$, where $e_i \neq e_\varnothing$, the pseudo embedding $h_{m_i}$ (Equation~\ref{eq:pseudo}) should be close to the entity embedding of the annotated entity, $\mathtt{EntEmbed}(e_{m_i})$. 
\begin{align*}
\mathtt{EL Loss}&=  \sum_{m_i}\alpha_i \cdot \mathbbm{1}_{{e}_{m_i} \neq e_{\varnothing}}\\
\alpha_i &= \frac{\exp(\mathtt{EntEmbed}(e_{m_i}) \cdot h_{m_i})}{\sum_{e \in \mathcal{E}}{\exp(\mathtt{EntEmbed}(e) \cdot h_{m_i}})}
\end{align*}
Note that this loss is not applied to 88\% of mentions, which do not have hyperlinks. Memory access for those mentions is unsupervised.
The same loss is used for the \texttt{EntityPred} head.

\paragraph{Masked Language Modelling}
We follow \bert, and train the \texttt{TokenPred} head to independently predict each of the masked out tokens in an input context, using a cross-entropy loss over $\mathcal{V}$.

%% file: background.tex
\section{Related Work}\label{sec:background}
\paragraph{Knowledge-augmented language models}

Our work shares inspiration with other approaches seeking to inject knowledge into language models \cite{ahn2016neural, yang2016reference, logan2019barack, ernie, sensebert, pretrained_encyc, knowbert, ebert, kadapter}. These have used a variety of knowledge sources (WikiData, WordNet relations, outputs from dependency parsers) and additional training objectives (synonym and hyponym-hypernym prediction, word-supersense prediction, replaced entity detection, predication prediction, dependency relation prediction, entity linking).\footnote{See also Table 1 of \citet{kadapter} for a useful review of such approaches.} Our focus is on adding knowledge about entities, so our work is closer to \citet{ernie, knowbert, wklm, kadapter, ebert} than to the linguistically-augmented approaches of \citet{sensebert, libert}.
Closest to our work, \knowbert \citep{knowbert} introduce an entity memory layer that is similar to the one in \ourapproach{}. In contrast with our work, \knowbert starts from the \bert~checkpoint, does not train with a knowledge-focused objective such as our mention-masking input function and uses pre-computed entity representations when integrating the information from knowledge bases. In addition, \knowbert relies on a fixed, pre-existing candidate detector (alias table) to identify potential candidates and entities for a span while our model learns to detect mentions. We compare to their approach in Section~\ref{sec:embedding_comp}.

\paragraph{Memory Augmented Neural Networks}

Our entity memory layer is closely tied to memory-based neural layers \cite{memory_networks, e2e_memory_networks}. In particular, it can be seen as a memory network where memory access is supervised through entity linking, and memory slots each correspond to a learned entity representation. When unconstrained, these memory networks can be computationally expensive and supervising access through entity linking limits this issue. Another approach to scale memory networks is given by \citet{largepkm} who introduce product-key memories to efficiently index a large store of values. 

\paragraph{Conditional Computation}

Conditional computation models seek to increase model capacity without a proportional increase in computational complexity. This is usually done through routing, where only a subset of the network is used to process each input. To facilitate such routing, approaches such as large mixture of experts \cite{mixtureofexperts} or gating \cite{eigen_gating,cho_gating} have been used.
Our method proposes entities as experts, which allows us to supervise memory access at two levels.
We only access memories for entity mentions, and we only need to access memories for the entities that were mentioned.

%% file: our_models.tex
\section{Models Evaluated}\label{sec:models}
We evaluate \ourapproach{} on cloze knowledge probes, open-domain question answering and relation extraction.  Here, we describe baselines from previous work and ablations.

\subsection{The Entities as Experts models}\label{sec:eae_model_description}

Our primary model, {\bf E}ntities {\bf A}s {\bf E}xperts embeds the input using the token embedding layer, then passes it through 4 transformer layers. This output is used for the entity memory layer, which uses embeddings of size 256, then passed through an additional 8 transformer layers. Finally, the model has a \texttt{TokenPred} head and an \texttt{EntityPred} head.
In \ourapproach{}, hyper-parameters for the transformer layers are identical to those in BERT-base \cite{bert}. 

\paragraph{Sparse Activation in EaE}
\ourapproach{} only accesses the entity embedding matrix for mention spans, and we only retrieve $k=100$ entity memories for each mention (see Appendix~\ref{sec:top_k} for analysis of this choice). 
This type of conditional computation can facilitate massive model scaling with fixed computational resources
\cite{mixtureofexperts,largepkm} and, while our current implementation does not yet include an efficient implementation of top-k routing (Section~\ref{sec:model}), we note that it is possible with fast maximum inner product search \cite{ram2012maximum,shen2015learning, shrivastava2014asymmetric, FAISS}. We leave the implementation and investigation of this to future work.

\subsection{Ablations}

\paragraph{EaE-unsup}
In \ourapproach{}-unsup, the entity memory is not supervised to isolate the usefulness of supervising memory slots with entity linking. We use full attention at inference when doing memory access.

\paragraph{No-EaE}
This ablation seeks to isolate the impact of the entity-memory layer. No~\ourapproach{} has a token embedding layer, twelve transformer layers, and an \texttt{EntityPred} and a \texttt{TokenPred} head. This approach has similar number of parameters as \ourapproach{},\footnote{With the exception of projection matrices totalling less than one million parameters.} but only uses the entity embedding matrix at the \texttt{EntityPred} head.
In contrast with \ourapproach{}, this baseline cannot model interactions between the entity representations in the entity embedding matrix. Also, the entity embeddings cannot be directly used to inform masked language modelling predictions.

\paragraph{BERT / MM} We compare to the \bert{} model. To ascertain which changes are due to \ourapproach{}'s data and masking function (Section~\ref{sec:objective}), and which are due to the entity-specific modeling, we report performance for \bertmm{} which uses \bert's architecture with \ourapproach{}'s masking function and data. We present results for Base and Large model sizes.

\subsection{Question Answering Models}\label{sec:previous_work_baselines}

\relic{} learns entity embeddings that match \bert{}'s encoding of the contexts in which those entities were mentioned \cite{relic}. 
\tfive{} is an encoder-decoder trained on an enormous web corpus. We compare to the version fine-tuned for open-domain question answering \cite{t5_openqa}.
We also compare to the nearest neighbour baselines introduced by \citealt{lewis2020question}; and we compare to three recent QA models that use a retriever and a reader to extract answers: 
BM25+\bert{} and {\sc Orqa} from \citealt{orqa} and GraphRetriever (GR) is introduced by \citealt{min_graphretriever}.
All are described fully in Appendix~\ref{sec:models_compared}.

%% file: knowledge_probing_tasks_and_results.tex
\section{Knowledge Probing Tasks}\label{sec:tasks}

We follow previous work in using cloze tests and question answering tasks to quantify the declarative knowledge captured in the parameters of our model \cite{language_as_kb, t5_openqa}. 

\subsection{Predicting Wikipedia Hyperlinks}\label{sec:intrinsic_task}
We explore the ability of our model to predict masked out hyperlink mentions from Wikipedia, similar to the pre-training task\footnote{In pre-training, we also mask non-hyperlinked mentions.} (Section~\ref{sec:training}). 
We calculate accuracy on a 32k test examples separate from the training data (Appendix~\ref{sec:pretraining}).

Table~\ref{tab:intrinsic_compare} shows the results for all our models.
The MM-base and No-\ourapproach{} models perform similarly on the token prediction task. 
These two models have the same architecture up until the point of token prediction.
This indicates that the signal coming from the entity linking loss (Section~\ref{sec:objective}) does not benefit language modeling when it is applied at the top of the transformer stack only.

Introducing the entity memory layer in the middle of the transformer stack (\ourapproach{}) improves performance on both language modeling and entity linking, compared to the No-\ourapproach{} model. 
This indicates that the entity representations are being used effectively, and that the model is learning inter-entity relations, using the output from the entity memory layer to improve predictions at the downstream entity and token prediction layers.

If the memory layer is not supervised (\ourapproach-unsup), performance on token prediction accuracy and perplexity is significantly worse than for \ourapproach{}.
This underlines the importance of entity linking supervision in teaching \ourapproach{} how to best allocate the parameters in the entity memory layer.

Finally, \ourapproach{} performs better at predicting mention tokens than the 24-layer MM-large, but does marginally worse in terms of perplexity. 
We believe that \ourapproach{} is overconfident in its token predictions when wrong, and we leave investigation of this phenomenon to future work. 

\begin{table}[tbp]
    \small
    \centering
    \begin{tabular}{l|cccc}
\toprule
         &        &            & \multicolumn{2}{c}{Token} \\
Model    & Params & Entity Acc & Acc & PPL \\ 
\midrule
MM-base  &  110m  & -          & 45.0 & 19.6\\
MM-large &  340m  & -          & 53.4 & \textbf{10.3} \\
\ourapproach{}-unsup & 366m  & -          & 46.9 & 16.9 \\     
No \ourapproach{}   &  366m  & 58.6       & 45.0 & 19.3 \\
\ourapproach{}     &  367m  & \textbf{61.8}       & \textbf{56.9} & 11.0 \\
\bottomrule
    \end{tabular}
    \caption{Results on cloze-style entity prediction accuracy, token prediction accuracy and perplexity on the test set of our masked hyperlink data.}
    \label{tab:intrinsic_compare}
\end{table}

\subsection{LAMA}\label{sec:lama_task}
\begin{table}[tb]
    \centering
    \small
    \begin{tabular}{l|gccc|c}
    \toprule
    Model &  Concept & RE & SQuAD & T & Avg. \\
    & Net &&&-REx&\\
    \midrule
    BERT-base   & 15.6 & 9.8 & 14.1 & 31.1 & 17.7 \\
    BERT-large  & \textbf{19.2} & \textbf{10.5} & 17.4 & 32.3 & 19.9 \\ 
    \hline
    MM-base  & 10.4 & 9.2 & 16.0 & 29.7 & 16.3 \\
    MM-large & 12.4 & 6.5 & \textbf{24.4} & 31.4 & 18.7 \\
    \ourapproach{}-unsup & 10.6 & 8.4 & 23.1 & 30.0 & 18.0 \\
    No \ourapproach{}    &  10.3 & 9.2 & 18.5 & 31.8 & 17.4 \\
    \ourapproach{}       &  10.7 & 9.4 & 22.4 & \textbf{37.4} & \textbf{20.0} \\
    \bottomrule
    \end{tabular}
    \caption{Results on the \lama{} probe.
    Adding entity memory improves performance for the probes that focus on entities. Mention masking reduces performance on ConceptNet sub-task which requires prediction of non-mention terms such as ``happy''.}
    \label{tab:lama}
\end{table}
\lama{}~\cite{language_as_kb} contains cloze tasks from three different knowledge base sources, and one QA dataset. \lama{} aims to probe the knowledge contained in a language model, with a focus on the type of knowledge that has traditionally been manually encoded in knowledge bases.
As a zero-shot probing task, it does not involve any task specific model training, and we do not apply any \lama{}-specific modeling.
Table~\ref{tab:lama} shows that adding the entity memory layer to the MM-base model improves performance across the board on this task.
Not supervising this layer with entity linking (\ourapproach{}-unsup) is worse overall. 

\ourapproach{}'s average accuracy is similar to BERT-large.
However, the \lama{} sub-task accuracies show that the two models are complementary. 
Mention focused approaches are much better than the BERT baselines at predicting the mention like words in the SQuAD and T-REx probes, but they are marginally worse for the RE probe and very significantly worse for the ConceptNet probe.
This is because the ConceptNet sub-task mostly includes non-entity answers such as ``fly'', ``cry'', and ``happy'', and a third of the answers in the Google-RE sub-task are dates.
We leave modeling of non-entity concepts and dates to future work.
Finally, the entity specific memory in \ourapproach{} is most beneficial for T-Rex, which focuses on common entities that are likely in our 1m entity vocabulary.

\subsection{Open domain Question Answering}\label{sec:opendomain_qa}
\paragraph{Setup}
\triviaqa{} and WebQuestions were introduced as reading comprehension and semantic parsing tasks, respectively \cite{triviaqa, webquestions}.
More recently, these datasets have been used to assess the performance of QA systems in the open domain setting where no evidence documents or database is given. In this setup, TriviaQA contains 79k train examples and WebQuestions 3.1k. 
Most approaches rely on a text corpus at test time, extracting answers from evidence passages returned by a retrieval system.
However, \tfive{} and \relic{} used neural networks to answer questions directly.
We follow \citealt{t5_openqa} in describing approaches as open-book (test time access to corpus) and closed-book (no test time access to corpus), and we report the nearest neighbour results from \citealt{lewis2020question}.%

We follow \relic{} and resolve answer strings to Wikipedia entity identifiers.\footnote{We resolve 77\% of TriviaQA train; 84\% of TriviaQA dev; 84\% of WebQuestions train; 95\% of WebQuestions dev; 91\% of WebQuestions test. See Appendix~\ref{sec:open_domain_qa_appendix} for full details. Answering with free-form text is out of this paper's scope} We follow the training procedure from Section~\ref{sec:training}, with a round of task specific training that applies the entity linking and mention detection losses to the question answering data. 
Each question is appended with a special `answer position' token and \ourapproach{} is trained to predict the correct answer entity in this position, using the entity linking loss from Section~\ref{sec:objective}.
Mention spans are identified within the question (Section~\ref{sec:data}) and the mention detection loss from Section~\ref{sec:objective} is applied to encourage \ourapproach{} to access the entity memory for entities in the question. See Appendix~\ref{sec:open_domain_qa_appendix} for additional information on task setup and fine-tuning hyper-parameters.

\begin{table}
    \centering \footnotesize
    \begin{tabular}{lcccc}
\toprule
  & \#  & TQA & TQA       & Web \\
&Params & Dev & Wiki Test & Q \\
\midrule
\multicolumn{5}{c}{\emph{Open-Book}: Span Selection - Oracle 100\%}\\
\midrule
BM25+BERT & 110m & 47.1 & - & 17.7 \\
\textsc{Orqa}  & 330m & 45.0 & - & 36.4\\
GR  & 110m & 55.4 & - & 31.6 \\ 
\midrule
\multicolumn{5}{c}{\emph{Closed-Book}: Nearest Neighbor} \\
\midrule
\sc{Oracle} & -   &  63.6 & - & -  \\
TFIDF       & -    & 23.5 & - & -  \\ 
BERT-base   & 110m & 28.9 & - & -  \\
\midrule
\multicolumn{5}{c}{\emph{Closed-Book}: Generation - Oracle 100\% }\\ 
\midrule
T5-Base & 220m & - & 29.1 & 29.1\\
T5-Large~ &  770m & - & 35.9 & 32.2\\
T5-3B & 3B & - & 43.4 & 34.4 \\
T5-11B & 11B & 42.3 & 50.1 & 37.4 \\
T5-11B+SSM & 11B & 53.3 & 61.6 & 43.5 \\

\midrule
\multicolumn{5}{c}{\emph{Closed-Book}: Entity Prediction}\\ 
\midrule
{\sc Oracle}      & -    & 85.0 & - & 91.0 \\
\textsc{Relic}    & 3B   & 35.7 & - & -    \\
No \ourapproach{} & 366m & 37.7 & - & 33.4 \\
\ourapproach{}    & 367m & 43.2 & 53.4 & 39.0 \\
\ourapproach{}, emb 512 & 623m & 45.7 & - & 38.7 \\
\bottomrule
    \end{tabular}
    \caption{Exact Match accuracy on TriviaQA and Web\-Questions.
    Open-book approaches reserve 10\% of the training data for development; entity prediction approaches only train on linked entity answers; and \tfive{} merges Unfiltered-Dev into training for Wiki-Test. For more description of these choices, see Appendix~\ref{sec:tqa_detail_eval}.}
    \vspace{-5pt}
    \label{tab:triviaqa_compare}
\end{table}

\paragraph{Results}
Table~\ref{tab:triviaqa_compare} shows results on two open domain QA datasets. Entity prediction methods, \relic{} and \ourapproach{}, significantly outperform nearest neighbor baselines, showing that model generalizes beyond train / development overlap and entity representations contains information about answer entities. 
No-\ourapproach{} and \relic{} both encode text with a transformer and retrieve answer entities.
Compared to \relic{}, No-\ourapproach{} is trained to identify all entities in a piece of text, instead of just one. This leads to small but significant gains on TriviaQA.
A much larger gain (almost 6 points on both datasets) comes from adding an entity memory inside the transformer encoder (\ourapproach{}).
We also show that it is possible to improve performance on TriviaQA by doubling the size of the entity embeddings to 512d (\ourapproach{}-emb-512).
While this almost doubles the model size, it does not significantly increase the number of parameters that need to be accessed at inference time.
See Section~\ref{sec:eae_model_description} for a discussion of how this could be beneficial, and Appendix~\ref{sec:top_k} for a preliminary investigation of conditional activation in \ourapproach{}.

Even though entity prediction approach cannot answer $15\%$ of the data with unlinked answers for TriviaQA, and 9\% for WebQuestions, it outperforms all of the standard \tfive{} models including one that has $30\times$ the parameters.
This indicates that the entity specific model architecture is more efficient in capturing the sort of information required for this knowledge probing task than the general encoder-decoder architecture used by \tfive{}.
However, when \tfive{} is enhanced with an extra pre-training steps focusing on likely answer spans from Wikipedia (\tfive{}-11B + SSM) its performance leapfrogs that of \ourapproach{}.
We note that the `salient spans' included in the SSM objective are likely to be entities \cite{realm}, and believe that there is significant future work to be done in combining methods of entity prediction and text generation.

Though closed-book approaches are still behind open-book approaches on TriviaQA, we believe even higher performances could be attained by ensembling diverse approaches and a preliminary study (Appendix~\ref{sec:paradigms}), indicates that ensembling open-book with closed-book approaches is preferable to ensembling within a single paradigm.

%% file: analysis.tex
\begin{figure*}
    \begin{subfigure}{0.33\textwidth}
    \includegraphics[width=0.9\textwidth]{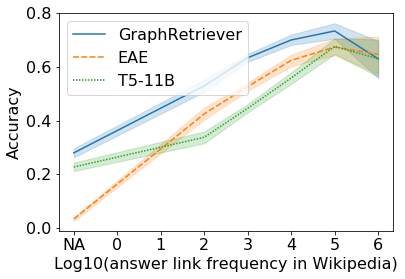}
    \subcaption{}
    \end{subfigure}
    \begin{subfigure}{0.33\textwidth}
    \includegraphics[width=0.9\textwidth]{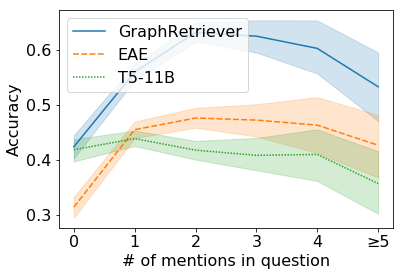}
    \subcaption{}
    \end{subfigure}
    \begin{subfigure}{0.33\textwidth}
    \includegraphics[width=0.9\textwidth]{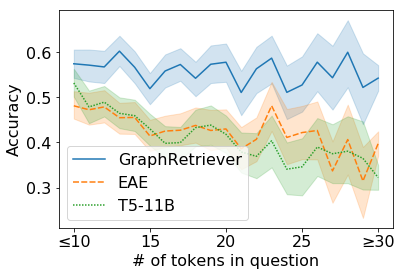}
    \subcaption{}
    \end{subfigure}
    \caption{Performance on TriviaQA by: answer frequency in our Wikipedia training corpus (NA if not linked);      proper names in the question; tokens in the question. Standard deviations obtained through bootstrapping.}
    \label{fig:quantitative_analysis}
\end{figure*}

\section{Analysis of TriviaQA Results}

\subsection{Entity-Based Analysis}
We compare the performances of retrieval-based \textsc{GR}, generation-based {\tfive{}-11B,} and our {\ourapproach{} model} on the TriviaQA Unfiltered-dev set. 
Figure~\ref{fig:quantitative_analysis} (a) shows that all models perform better on frequent entities. \ourapproach{}'s performance is much lower for entities seen fewer than one hundred times, likely because their entity embedding do not contain enough information.
Figure~\ref{fig:quantitative_analysis} (b) shows that as the number of named entity mentions grows in the question, T5 performance decreases whereas \ourapproach{} and \textsc{GR} performance increases. We presume more entity mentions makes it easier to retrieve relevant documents, thus contributing to better performance for \textsc{GR}. For \ourapproach{}, having more entity mentions allows the model to access more entity knowledge. We conduct further qualitative analysis below.
Figure~\ref{fig:quantitative_analysis} (c) shows that closed-book models under-perform on long questions. The trend disappears for open-book models.

\subsection{Manual Analysis}
We randomly sampled 100 examples from unfiltered-dev set, and analysed the ability of \ourapproach{}'s to correctly identify and link the question's entities. We find that $87\%$ of questions do not have any incorrectly predicted entities.\footnote{We do not define a strict bracketing to decide which entities in nested phrases like $\rm[[1966~[FIFA~World~Cup]]~Final]$ should be predicted.}

\input{triviaqa_analysis_examples_small}
We find that when there is at least one incorrectly linked entity, the performance of \ourapproach{}~is considerably reduced. On this small sample, the performance is even lower than for examples in which there are no named entities in the question. 

Table~\ref{fig:tqa_examples} illustrates three representative examples from \ourapproach{} and T5.
In the first example, the question contains a date but no proper names.
Since \ourapproach{} does not have a representation for dates in the entity memory, this question is challenging and the model predicts an incorrect answer of the correct type (annual event).
The second example demonstrates \ourapproach{}'s ability to model connections between entities. 
In this case, `The Master' only occurs 38 times in our training data, but in each of those occurrences it is likely that at least one of the relevant actors is also mentioned. \ourapproach{} learns the character-actor relationship, while T5 makes up an incorrect character name based on a common category of Dr. Who villain.
The final example highlights the sensitivity of \ourapproach{} to incorrectly linked question entities. 
Here, the name `Jason' has been incorrectly linked to the Greek mythological with that name, which causes \ourapproach{} to predict another Greek mythological figure, Icarus, as the answer.
This is particularly interesting because Icarus is strongly associated with human flight---\ourapproach{} is still trying to find an aviator, albeit one from Greek mythology. Additional examples can be found in Appendix~\ref{sec:additional_triviaqa_examples}.

\subsection{Top-K over Entity Embeddings}\label{sec:top_k}

\begin{table}
\small
    \centering
    \begin{tabular}{l|ccc|c}
\toprule
$K$ & Entity acc & Tok acc & Tok PPL & TQA \\
\midrule
1 & 59.2 & 56.7 & 18.0 & 40.1  \\ 
10 & 61.7 & 57.2 & 11.1 & 43.1 \\ 
100 & 61.8 & 57.1 & 11.0 & 43.2 \\
Full ($10^6$) & 61.8 & 56.9 & 11.0 & 43.4\\ \bottomrule
    \end{tabular}
    \caption{Impact of varying the number of retrieved entity embeddings ($K$) in the Entity Memory layer at inference on the entity prediction and TriviaQA tasks.}
    \label{tab:topk_analysis}
\end{table}

As described in Section~\ref{sec:model}, \ourapproach{} uses the top 100 entity memories during retrieval for each mention. Here, we empirically analyse the influence of this choice.
Table~\ref{tab:topk_analysis} shows how varying the number of retrieved entity embeddings in the entity memory layer at inference time impacts accuracy of entity prediction and TrviaQA.  Even for $K=10$, performance does not deteriorate meaningfully. 

This is a key advantage of our modular approach, where entity information is organized in the Entity Memory layer. Despite only accessing as many parameters as \bert{} Base, our model outperforms \bert{} Large on token prediction. Similarly in Trivia\-QA, we outperform T5-3B model, while accessing about 3\% of the parameters. While naive implementation would not bring significant computational gain, fast nearest neighbor methods methods on entity embeddings enable retrieving the top $K$ entities in sub-linear time and storage, enabling efficient model deployment.

%% file: triviaqa_analysis_examples_small.tex
\begin{table*}[!ht]
\footnotesize
\begin{tabular}{p{0.43\textwidth}p{0.16\textwidth}p{0.16\textwidth}p{0.15\textwidth}}\toprule
\textbf{Question}  & \textbf{Answer}        & \textbf{T5 Prediction} & \textbf{EaE Prediction} \\
\midrule
Next Sunday, Sept 19, is International what day? & Talk like a pirate day & talk like a pirate day & Pearl Harbor Remembrance Day \\
 \midrule
Which \textcolor{ForestGreen}{\bf [Dr. Who]} villain has been played by \textcolor{ForestGreen}{\bf[Roger Delgado]},  \textcolor{ForestGreen}{\bf[Anthony Ainley]},  \textcolor{ForestGreen}{\bf[Eric Roberts]}, etc? &  The Master & mr. daleks & The Master  \\
\midrule
Which early aviator flew in a plane christened {\textcolor{red}{\bf \{Jason\}}}? & Amy Johnston & jean batten & Icarus \\ 
\cmidrule{1-1} 
${\rm Jason}~\rightarrow~{\rm Jason~(Greek~Mythology)~Q176758}$ \\
\bottomrule
\end{tabular}
\caption{Illustrative examples of predictions for the TriviaQA dev set. Questions are annotated with \textcolor{ForestGreen}{\bf [correct]} and \textcolor{red}{\bf \{incorrect\}} entity predictions from \ourapproach{}, which is most successful when question entities are linked successfully.}
\label{fig:tqa_examples}
\end{table*}

%% file: entity_rep_comparison.tex
\section{Comparison to Alternative Entity Representations}\label{sec:embedding_comp}

\begin{table*}
    \small
    \centering
    \begin{tabular}{lcc|c|cc|cc|c}
    \toprule
Initialization & Dim  & Learned & Typing &  PPL      & Entity Acc. & LAMA-SquAD & T-REx & TQA \\
 \midrule
Random         &  100 & \cmark{} & 74.5  & 10.6      & 63.0        & 17.7       & 32.4  & 30.9 \\
TransE         &  100 & \xmark{} & 74.1  & 14.7      & 49.7        & 12.8       & 30.8  & 24.2 \\
TransE         &  100 & \cmark{} & \textbf{75.1}  & 10.5      & 63.1        & 16.3       & 31.9  & 30.6 \\ 
Random         &  300 & \cmark{} & 74.2  & 9.4       & 64.5        & \textbf{19.3}       & 31.7  & 33.2 \\
Deep-Ed        &  300 & \xmark{} & 73.0  & 12.6      & 57.4        & 17.0       & 29.8  & 30.1 \\
Deep-Ed        &  300 & \cmark{} & 74.2  & {\bf 8.9} & {\bf 65.1}  & 15.7      & \textbf{33.5}  & \textbf{33.9} \\
 \bottomrule
    \end{tabular}\vspace{-5pt}
    \caption{Comparison of learned representations to knowledge graph embeddings (TransE) and pre-trained representations of entity descriptions and contexts (Deep-Ed). All experiments use same 200k entities.}
    \label{tab:embed_comparison}
\end{table*}

\ourapproach{} is novel in learning entity representations along with the parameters of a Transformer model. Here, we provide a direct comparison between \ourapproach{}'s learned representations and the externally trained, fixed representations used by \ernie{} \cite{ernie} and \knowbert{} \cite{knowbert}.
Table~\ref{tab:embed_comparison} presents results for the \ourapproach{} architecture trained with entity memories initialized from three different sources and two different training scenarios.
The initial embedddings are either the TransE-Wikidata embeddings used by \ernie{} \cite{transe}; the Deep-Ed embeddings used by \knowbert{} \cite{ganea-hofmann-2017-deep}; or the random embeddings used by \ourapproach{}.
The embeddings are either frozen, following \ernie{} and \knowbert{}, or trained along with all other network parameters.\footnote{All experiments use the same 200k entities, at the intersection of the sets used by \ernie{} and \knowbert{} For efficiency, these three comparison systems are trained for 500k steps, rather than the full 1m steps used in Section~\ref{sec:tasks}.}
Along with the knowledge probing tasks from Section~\ref{sec:tasks}, we report performance on the on the 9-way entity typing task from \citealt{ufet} (Appendix~\ref{sec:typing}).%

It is clear that learning entity representations is beneficial.
However, initializing the entity memory with Deep-Ed embeddings leads to gains on most of the knowledge probing tasks, suggesting that there are potential benefits from combining training regimens. Meanwhile, the best entity typing results come from initializing with Wikidata embeddings, possibly because Wikidata has high quality coverage of the types (`person', `organization', `location', etc.) used.
Finally, we point out that both \ernie{} and \knowbert{} differ from \ourapproach{} in other ways (see Appendix~\ref{sec:models_compared}). 
\knowbert{} in particular incorporates WordNet synset embeddings as well as entity embeddings, leading to entity typing accuracy of 76.1.
Future work will explore different ways of combining learned embeddings with knowledge from diverse external sources.

As discussed in Section~\ref{sec:background}, there are significant differences between \ourapproach{} and \knowbert{} other than the choice of entity representation. 
In particular, \knowbert{} has an explicit entity-entity attention mechanism.
To determine whether this has a significant effect on a model's ability to model entity-entity relations, we evaluate \ourapproach{} on TACRED \cite{tacred} using the cleaned dataset introduced by \cite{tacred-revisited}.\footnote{Our method follows~\cite{matching-the-blanks} in using special tokens to mark subject and object, and concatenating their representations to model the relation.}
Table~\ref{tab:tacred} shows that \ourapproach{} outperforms \knowbert{} on the revised and weighted splits introduced by \cite{tacred-revisited}, although it slightly under-performs on the original setting.\footnote{The weighted split weights examples by its difficulty, focusing on correctly predicting difficult examples.}
This result indicates that \ourapproach{}, without explicitly entity-entity attention, can capture relations between entities effectively.

\begingroup
\setlength{\tabcolsep}{5pt} %
\begin{table}[tbp]
    \small
    \centering
    \begin{tabular}{l|cccc}
\toprule
Model          & \# Params   & Original & Revised & Weighted \\ 
\midrule
\knowbert   &  523m  & \textbf{71.5} & 79.3 & 58.4  \\
\ourapproach{} &  366m  & 70.2 & \textbf{80.6} & \textbf{61.3}  \\     
\bottomrule
    \end{tabular}
    \caption{F$_{1}$ scores on original and revisited versions of TACRED test sets. \knowbert scores are reported as in~\citet{tacred-revisited}, corresponding to the best performing variant (\texttt{KnowBert-W+W}).}
    \label{tab:tacred}
\end{table}
\endgroup

%% file: conclusion.tex
\section{Conclusion}
We introduced a new transformer architecture, \ourapproach, which learns entity representations from text along with other model parameters. Our evaluation shows that \ourapproach{} is effective at capturing declarative knowledge and can be used for a wide variety of tasks -- including open domain question answering, relation extraction, entity typing and knowledge probing tasks. Our entity representations influence the answer predictions for open-domain question answering system and are of high quality, compared to prior work such as \knowbert.

Our model learns representations for a pre-fixed vocabulary of entities, and cannot handle unseen entities. Future work can explore representations for rare or unseen entities, as well as developing less memory-intensive ways to learn and integrate entity representations. Furthermore, integrating information from knowledge-bases can further improve the quality of entity representation.

%% file: appendix.tex
\input{experimental_setup.tex}

\section{TriviaQA Evaluation Data Configuration}\label{sec:tqa_detail_eval}
As the TriviaQA dataset was originally introduced for reading comprehension task, prior work adapted the initial splits and evaluation to reflect the open domain setting. We describe the setup used by prior approaches and introduce ours, to enable fair comparison. The dataset comes in two blends, Wikipedia and Web documents. While TriviaQA's official web evaluation uses (question, document, answer) triplets, all open domain approaches average over (question, answer) pairs when using the Web data.

\paragraph{Open-book Approaches:}
\citet{orqa, min_hardem, min_graphretriever} use only the web splits. They use 90\% of the original train data for training, performing model selection on the remaining 10\% and reporting test numbers on the original development set since the original test set is hidden.

\paragraph{Previous Closed-book Approaches:}
\citet{t5_openqa} also use the web data to train and validate their model. However, they use the Wikipedia\footnote{\url{https://competitions.codalab.org/competitions/17208}} test split as a test set. After hyper-parameter tuning, they re-train a model on both the training and development sets of the web data. \citet{relic} uses the original web splits and reports performance on the development set.

\paragraph{Our approach:}
To compare our approach more closely to \tfive{}, we follow their setup, with the exception that we do not re-train a model on both the train and development splits after hyper-parameter selection.

\section{Comparing QA paradigms}\label{sec:paradigms}

\begin{table}[tbp]
\small
    \centering
    \begin{tabular}{l|cc}
\toprule
Systems                    & Oracle Acc.   & Pred Overlap (\%)  \\
\midrule
\tfive{} \& \ourapproach{} & 55.9          & 29.3 \\ 
\tfive{} \& GR             & \textbf{66.4} & 30.1 \\ 
\ourapproach{} \& GR       & 64.6          & 33.6\\
\textsc{Orqa} \& GR        & 63.8          & \textbf{39.6}\\ \bottomrule
    \end{tabular}
    \caption{Comparing prediction overlap and oracle accuracy on TriviaQA. Oracle accuracy considers a prediction correct if at least one of the model prediction is correct. While \textsc{Orqa} and \textsc{GR} outperform \ourapproach{} and \tfive{}, their predictions overlap more with \textsc{GR} and offer less complementary value.}
    \label{tab:compare_paradigms}
\end{table}

We compare the performance of four systems (\textsc{Orqa}, GraphRetriever (\textsc{GR}), \tfive{}, \ourapproach{}) on the TriviaQA Unfiltered-Dev set. \textsc{GR} achieves an accuracy of 55.4, \textsc{Orqa} 45.1, \ourapproach{} 43.2 and \tfive{} 42.3. As the open-book paradigm differs significantly from the closed-book one, we intuit they might complement each other.

To test this hypothesis, we measure prediction overlap and oracle accuracy. Oracle accuracy considers a prediction correct if either system is correct (see Table~\ref{tab:compare_paradigms}). Unsurprisingly, the two open book approaches show the most similar predictions, overlapping in nearly 40\% of examples. While \textsc{Orqa} outperforms \tfive{} and \ourapproach{}, the oracle accuracy of \textsc{Orqa} \& GR is lower than \textsc{Orqa} \& \tfive{} or \textsc{Orqa} \& \ourapproach{}. This suggests some questions might be better suited to the closed book paradigm. In addition, the oracle accuracy of the two closed book systems is higher than that of the best performing open book system. We leave designing approaches that better combine these paradigms to future work.

\section{Additional Examples of TriviaQA Predictions}\label{sec:additional_triviaqa_examples}

\input{additional_triviaqa_examples.tex}

%% file: experimental_setup.tex
\input{models_to_compare.tex}

\section{Wikipedia Pre-training}\label{sec:pretraining}

\paragraph{Wikipedia Processing} We build our training corpus of contexts paired with entity mention labels from the 2019-04-14 dump of English Wikipedia.
We first divide each article into chunks of 500 bytes, resulting in a corpus of 32 million contexts with over 17 million entity mentions. 
We restrict ourselves to the one million most frequent entities (86\% of the linked mentions).  These are processed with the BERT tokenizer using the lowercase vocabulary, limited to 128 word-piece tokens.
In addition to the Wikipedia links, we annotate each sentence with unlinked mention spans using the mention detector from Section~\ref{sec:training}. These are used as additional signal for our mention detection component and allow the model to perform retrieval even for mentions that are not linked in Wikipedia. 
We set aside 0.2\% of the data for development and 0.2\% for test and use the rest to pre-train our model.

\paragraph{Pre-training hyper-parameters} We pre-train our model from scratch. We use \textsc{Adam} \cite{adam} with a learning rate of 1e-4. We apply warmup for the first 5\% of training, decaying the learning rate afterwards. We also apply gradient clipping with a norm of 1.0. We train for one million steps using a large batch size of 4096. We use a \texttt{TokenPrediction} head for all our models and an \texttt{EntityPrediction} head for the \ourapproach{} and No-\ourapproach{} models. We did not run extensive hyper-parameter tuning for pre-training due to computational cost. We train on 64 Google Cloud TPUs for all our pre-training experiments. All pre-training experiments took between 2 days and a week to complete.

\section{Open Domain Question Answering}\label{sec:open_domain_qa_appendix}

\paragraph{Open Domain QA Preprocessing} We annotate each question with proper-name mentions\footnote{Nominal mentions in questions typically refer to the answer entity. This is unlike Wikipedia text, where nominal mentions refer to entities that have previously been named. We only link proper name mentions in questions so that the model is not forced to hallucinate links for entities that have not been properly introduced to the discourse.} using the mention detector from Section~\ref{sec:training}.

When the answer is an entity, in our entity vocabulary, we link the answer string to an entity ID using the {\sc sling} phrase table\footnote{\url{https://github.com/google/sling}} \cite{ringgaard2017sling}. If the answer is not an entity in our vocabulary, we discard the question from the training set, though we keep it in the development and test set to ensure fair comparisons with prior work. Table~\ref{tab:linking_proportion} shows the share of answers that were linked using our procedure. This means that Oracle performance for our model on the TriviaQA development set is only 85\%, which is due to non entity-answers and entities not in our vocabulary.

\begin{table}[tbp]
    \small
    \centering
    \begin{tabular}{l|ccc}
\toprule
Dataset    & Train & Development & Test\\
\midrule
TriviaQA  & 77\% & 84\% & - \\
WebQuestions & 84\% & 95\% & 91\% \\
\bottomrule
    \end{tabular}
    \caption{Share of the answers that are linked by our linking procedure for the TriviaQA and WebQuestion datasets. The test set for TriviaQA is not public, hence the missing number.}
    \label{tab:linking_proportion}
\end{table}

\paragraph{Hyper-parameters} For TriviaQA, we fine-tune the entire model using a learning rate of 5e-6, a batch size of 64 and performing 50,000 training steps. For WebQuestions, we set the learning rate to 3e-5, the batch size to 32 and train for 700 steps. Fine-tuning was done on 4 Google Cloud TPUs. In both cases, we searched the learning rate over $\{5\times10^{-5}, 3\times10^{-5}, 10^{-5}, 5\times10^{-6}, 3\times10^{-6}\}$ and the batch size in $\{32, 64, 128\}$ and selected the model based on validation performance.

\section{Entity Typing}\label{sec:typing}

We describe the procedure to obtain the Typing results of Table~\ref{tab:embed_comparison}.

\paragraph{Open Entity Processing}

We use the Ultra-fine entity typing dataset introduced in \cite{ufet}. As is done in \citet{ernie, knowbert} we limit the task to the 9 generic types (`person', `group', `organization', `location', `entity', `time', `object', `event' and `place') and use the Micro-F1 metric. The dataset is comprised of 5994 examples, equally divided between train, development and test. We pass the sentences to our model without extra annotations for mention annotations or entity linking, relying instead on the model's own predictions. To predict a type for the span, we take the span representation and project it to the 9 classes.

\paragraph{Hyper-parameters} Since we have no mention boundary or linking information, we freeze the entity embeddings and the mention detection parameters in fine-tuning. We used a learning rate of 3e-5, a batch size of 32 and trained for 700 steps. We also used label smoothing with a value of 0.1. We searched the learning rate and the batch size between the same bounds as for the open-domain question answering tasks. For every model, we ran with five different seeds and selected the best model based on validation performance before running on test. We selected the threshold to compute F1 based on validation scores.

%% file: models_to_compare.tex
\section{Other Models Evaluated}\label{sec:models_compared}

In this section we describe the models mentioned in Section~\ref{sec:previous_work_baselines} and compared to in Section~\cref{sec:tasks,sec:embedding_comp}.

\subsection{Open-Book Question Answering Systems}

Open-Book Open Domain Question Answering Systems are usually comprised of two components: a retriever and a reader. The retriever reads a set of documents from a corpus or facts from a knowledge base. Top retrievals are then fed to the reader which predicts an answer, often through span selection.

\paragraph{BM25+BERT} \cite{orqa} uses the non-parametric BM25 as its retriever and a \bert-base reader to predict an answer through span selection.

\paragraph{ORQA} \cite{orqa} uses two \bert-base models to retrieve relevant passages and a \bert-base reader to predict the answer span.

\paragraph{Graph Retriever} \cite{min_graphretriever}'s retriever can perform Graph-based, Entity-based and Text-match retrieval. These different retrieval modalities are then fused in representations that serve as input to a span-selection reader.

\subsection{Other Models}

\paragraph{BERT} \cite{bert} is a transformer, pre-trained using masked language modelling. We report results for BERT-base, which has 110m parameters, and BERT-large, which has 340m parameters. The transformer architecture used by BERT-base is identical to the 12 transformer layers in \ourapproach{}. BERT-large uses a much larger transformer, and has a similar number of parameters overall to \ourapproach{}.

\paragraph{RELIC} 
\cite{relic} is a dual encoder with a BERT-base architecture that compares a representation of a mention to an entity representation. It is similar to our No-\ourapproach{} architecture. Its training is however different, as only linked mentions are masked and only one mention is masked at a time. In addition, \relic{} does not have mention detection or masked language modelling losses. Finally, it is also initialized with the \textsc{Bert} checkpoint whereas we train our models from scratch. 

\paragraph{T5}
\tfive{} is an encoder-decoder transformer introduced in \citet{t5}. It has been fine-tuned for open domain question answering in \citet{t5_openqa}. In that setup, the model is trained to generate the answer to a question without any context. \tfive{} does not explicitly model entities or have any form of memory. We compare to models of different sizes, from 220m parameters to 11B. `SSM' refers to salient span masking, indicating that prior to fine-tuning on open domain question answering the model was fine-tuned using salient span masking, which bears resemblances to our mention masking.

\paragraph{KnowBERT} \cite{knowbert}  \knowbert is a \bert-base transformer that embeds multiple knowledge bases to improve performance in a variety of tasks. The integration of this information is done through a Knowledge Attention and Recontextualization component, which can be seen as a small transformer that is run on the pooled mention representations of potential entities. \knowbert uses this layer to embed entity representations from Wikipedia as well as Wordnet graph information. In contrast with our work, \knowbert starts from the \bert checkpoint, does not train with a knowledge-focused objective such as our mention-masking input function and uses pre-existing representations when integrating the information from knowledge bases. In addition, \knowbert relies on a fixed, pre-existing candidate detector (alias table) to identify potential candidates and entities for a span while our model learns mention detection.

\paragraph{\ernie} \cite{ernie} \ernie is a \bert-base transformer that takes as additional input the list of entities in the sentence. Multi-head attention is performed on those entities before they are introduced in they are aggregated with the token representations. In addition to \bert's pre-training objective, \ernie also masks entities and trains the model to predict them. In contrast with both \knowbert and \ourapproach{}, \ernie takes entity linking as an input rather than learning to do it inside the model. In contrast with our approach, \ernie uses pre-existing entity representations that are fixed during training.

%% file: additional_triviaqa_examples.tex
\begin{table*}[ht]
\footnotesize
\begin{tabular}{p{0.01\textwidth}p{0.4\textwidth}p{0.15\textwidth}p{0.15\textwidth}p{0.15\textwidth}}\toprule
    & \textbf{Questions with no proper name mentions}  & \textbf{Answer}        & \textbf{T5 Prediction} & \textbf{EaE Prediction} \\
\midrule
i  & What links 1st January 1660 and 31st May 1669?  &  First and last entries in Samuel Pepys's diaries & they were the dates of the henry viii's last bath & Anglo-Dutch Wars \\
\midrule
ii & Which radioactive substance sometimes occurs naturally in spring water? & radon & radon & radon \\
\midrule
& \textbf{Questions with only \textcolor{ForestGreen}{[correctly]} linked entities}      \\\midrule
iii & Who directed the 2011 \textcolor{ForestGreen}{[Palme d'Or]} winning film \textcolor{ForestGreen}{`[The Tree Of Life]'}? & Terence Malick & ang lee &  Terrence Malick  \\\midrule

iv & Name the year: \textcolor{ForestGreen}{[Hirohito]} dies; The \textcolor{ForestGreen}{[Exxon Valdez]} runs aground; \textcolor{ForestGreen}{[San Francisco]} suffers its worst earthquake since 1906. & 1989 & 1989 & 1990s  \\
\midrule
     & \textbf{Questions with \textcolor{red}{\{incorrectly\}} linked entities}    \\
\midrule
v  & Which car manufacturer produces the \textcolor{red}{\bf \{Jimmy\}} model? & Suzuki & suzuki & Brixham \\
\cmidrule{2-2}
&  ${\rm Jimmy}~\rightarrow~{\rm Marcos~Engineering~Q1637323}$ \\
\midrule
vi & Where do you find the {\textcolor{red}{\bf \{Bridal Veil\}}}, \textcolor{ForestGreen}{\bf [American]}, and \textcolor{ForestGreen}{\bf [Horseshoe Falls]}? & Niagara falls & niagra Falls & Niagara Falls \\
\cmidrule{2-2} 
&  ${\rm Bridal~Veil}~\rightarrow~{\rm Veil~(Garment)~Q6497446}$ \\
\bottomrule
\end{tabular}
\caption{Additional examples of question, answer and model predictions for the TriviaQA Unfiltered dev set.}
\label{fig:extra_tqa_examples}
\end{table*}

Table~\ref{fig:extra_tqa_examples} illustrates additional representative sample of questions and predictions from \ourapproach{} and T5. 
We break this sample down into questions that contain no named entities, questions that contain only correctly linked named entities, and questions that contain incorrectly linked named entities.

Example (i) shows another case where our model fails to handle dates. While T5 also has no distinct representation for dates, its 11B parameters managed to memorize the esoteric connection between the phrases `Sept 19' and `talk like a pirate day'.
In the second example without named entities, the answer entity is sufficiently frequent in Wikipedia (top 50k entities), and \ourapproach{} seems to have learned its connection with the specific categorical information in the question. 

Example (iii) shows another case where the correctly linked entities enable \ourapproach{} to correctly answer the prompt while \tfive{} predicts a different director's name. Example (vi) shows how \ourapproach{} fails at predicting date answers, even when there is abundant information in the question, predicting the entity representing the 1990's.

Examples (v) and (vi) shows other failure modes when \ourapproach{} fails to correctly predict the entities in the question. In example (vi) the entity is not available, though this not cause the model to err, likely thanks to the two other correctly predicted question entities. Example's (vi) typo (from `Jimny' to `Jimmy') is particularly interesting: when fixed, \ourapproach{} links `Jimny' correctly and predicts the right answer.